\pdfoutput=1
\documentclass[11pt]{article}

\usepackage{acl}
\usepackage{multirow}
\usepackage[utf8]{inputenc}
\usepackage{microtype}
\usepackage{times}
\usepackage{latexsym}
\usepackage{booktabs}
\usepackage{xcolor}
\usepackage{algorithm}
\usepackage[noend]{algpseudocode}

\usepackage{amsmath}
\usepackage{hyperref}
\usepackage{tikz}
\usepackage{pgfplots}
\pgfplotsset{compat=1.16}
\usepackage{multirow}
\usepackage{multicol}
\usepackage{caption}
\usepackage{graphicx}
\usepackage{tabu}
\usepackage{makecell}
\usepackage{soul}
\usepackage{booktabs}
\usepackage{array}
\usepackage{xcolor}
\usepackage{ragged2e}
\usepackage{subcaption}
\usepackage{footmisc}
\usepackage{algorithm}
\usepackage[noend]{algpseudocode}
\usepackage{xpatch}
\usepackage[T1]{fontenc}
\usepackage[utf8]{inputenc}

\usepackage{microtype}

\title{Practice Makes a Solver Perfect: Data Augmentation for Math Word Problem Solvers}

\author{Vivek Kumar, Rishabh Maheshwary and Vikram Pudi\\
Data Sciences and Analytics Center, Kohli Center on Intelligent Systems \\
International Institute of Information Technology, Hyderabad, India \\ {\{\text{vivek.k, rishabh.maheshwary\}@research.iiit.ac.in, vikram@iiit.ac.in}}}

\begin{document}
\maketitle
\begin{abstract}
Existing Math Word Problem (MWP) solvers have achieved high accuracy on benchmark datasets. However, prior works have shown that such solvers do not generalize well and rely on superficial cues to achieve high performance. In this paper, we first conduct experiments to showcase that this behaviour is mainly associated with the limited size and diversity present in existing MWP datasets. Next, we propose several data augmentation techniques broadly categorized into \textit{Substitution} and \textit{Paraphrasing} based methods. By deploying these methods we increase the size of existing datasets by five folds. Extensive experiments on two benchmark datasets across three state-of-the-art MWP solvers shows that proposed methods increase the generalization and robustness of existing solvers. On average, proposed methods significantly increase the state-of-the-art results by over five percentage points on benchmark datasets. Further, the solvers trained on the augmented dataset performs comparatively better on the challenge test set. We also show the effectiveness of proposed techniques through ablation studies and verify the quality of augmented samples through human evaluation.
\end{abstract}

\section{Introduction}
A Math Word Problem (MWP) consists of natural language text which describes a world state involving some known and unknown quantities, followed by a question text to determine the unknown values. The task is to parse the problem statement and generate equations that can help find the value of unknown quantities. An example of a simple MWP is shown in Table \ref{table:1}.
In recent years, the challenge of solving MWP has gained much attention in the NLP community as it needs the development of commonsense multi step reasoning with numerical quantities. With the rise of deep learning, performance of math solvers has also increased significantly over the years~\cite{wang-etal-2017-deep,zhang-etal-2020-graph-tree}. However, recent analysis conducted in~\cite{kumar-etal-2021-adversarial-examples} and~\cite{patel2021nlp} show that these deep learning based solvers rely on shallow heuristics to solve vast majority of problems. They curated adversarial examples and SVAMP challenge set respectively to infer that MWP solvers $(1)$ do not understand the relationship between numbers and their associated entities, $(2)$ do not focus on the question text and $(3)$ ignore word order information. In this paper, we first conduct experiments to establish that the above drawbacks are due to the limited size and diversity of problems present in the existing MWP datasets. Next, we propose various augmentation methods to create diverse and large number of training examples to mitigate these shortcomings. Our methods are focused on: $(1)$ Increasing the number of problems in the existing datasets and $(2)$ enhancing the diversity of the problem set.

% \usepackage{booktabs}
\begin{table}[]
\centering
\small
{\renewcommand{\arraystretch}{1}
\begin{tabular}{@{}l@{}}
\toprule
\begin{tabular}[c]{@{}l@{}}\textbf{Original Problem}\\ \textcolor{red}{Problem:} Nancy grew 8 potatoes. Sandy grew 5 potatoes.\\ How many potatoes did they grow in total ?\\
\textcolor{red}{True Equation:} X = 8+5\end{tabular} \\ \midrule
\begin{tabular}[c]{@{}l@{}}\textbf{Paraphrasing Method}\\ \textcolor{red}{Problem:} \textcolor{blue}{How many potatoes did they grow in all} \textcolor{violet}{given that}\\ nancy grew 8 potatoes and sandy grew 5 potatoes.\\
\textcolor{red}{Equation Label:} X = 8+5\end{tabular} \\ \midrule
\begin{tabular}[c]{@{}l@{}}\textbf{Substitution Method}\\ \textcolor{red}{Problem:} \textcolor{violet}{Dwight} grew 8 potatoes. \textcolor{violet}{Juliette} grew 5 potatoes.\\ \textcolor{blue}{How many potatoes did they grow together ?}\\
\textcolor{red}{Equation Label:} X = 8+5\end{tabular}\\ \bottomrule
\end{tabular}}
\caption{A MWP and its augmentation examples generated by our methods with preserved equation labels. Blue and Violet colours denote the changes made after the primary stage and secondary stage respectively.}
\label{table:1}
\end{table}
Training deep neural models effectively requires large number of data points~\cite{longpre2020effective}. Constructing large datasets which are annotated, labeled and have MWPs of similar difficulty level is a very expensive and tedious task. To address these key challenges, we resort to data augmentation techniques.
Our motivation behind generating augmentations is that humans require sufficient practice to understand MWPs. Humans learn to solve MWPs by going through a variety of similar examples and slowly become capable enough to tackle variations of similar difficulty levels. We aim to generate augmentations such that sufficient linguistic variations of a similar problem are present in the dataset. These variations will make the solver more robust in tackling MWP, increase their reasoning ability and numerical understanding.

Data augmentation for MWPs is a challenging task as we need to preserve the equation labels while generating new samples~\cite{kumar-etal-2021-adversarial-examples}. The generated samples should be $(1)$ semantically similar to their original counterpart, $(2)$ must have the same numerical values and preserve relationship with their respective entities and $(3)$ should maintain the same sequence of events in the problem text. Existing augmentation methods~\cite{wei-zou-2019-eda} cannot be directly applied due to the above mentioned reasons. Our methods can be broadly classified as follows:
\begin{itemize}
    \item \textbf{Paraphrasing Methods:} It generates variations of the question text by re-statement such that the semantic and syntactic meaning along with the equation labels is preserved.
    \item \textbf{Substitution Methods:}
    These methods generate variations of the problem statement by identifying and substituting some of the keywords such that the augmentations are semantically and syntactically correct.
\end{itemize}
To ensure high quality augmentations \footnote{Codebase and augmented datasets are available at: \href{https://github.com/kevivk/MWP-Augmentation}{https://github.com/kevivk/MWP-Augmentation}}, we propose a selection algorithm which selects samples that have high similarity with original problem and incur high loss values when tested on existing solvers. This algorithm helps selecting only those samples that can make existing solvers more robust. Further, we also verify the validity and the quality of generated augmentations through human evaluation.

Most of the existing MWP datasets are either in languages other than English or contain problems of varying difficulty levels~\cite{koncel-kedziorski-etal-2016-mawps,wang-etal-2017-deep,huang-etal-2016-well,amini2019mathqa,miao-etal-2020-diverse}.
We focus on strengthening existing English language datasets which can facilitate the development of better MWP solvers. We consider datasets containing MWP that can be solved using linear equations in one variable. These datasets include MaWPS~\cite{koncel-kedziorski-etal-2016-mawps} and ASDiv-A~\cite{miao-etal-2020-diverse} both having $2,373$ and $1,213$ problems respectively.
Following are the key contributions made in this paper:
\begin{itemize}
    \item To the best of our knowledge, this is the first work that extensively evaluates data augmentation techniques for MWP solving. This is the first attempt to generate MWP problems automatically without manual intervention.
    \item Accuracy of the state of the art solvers increases after training on the proposed augmented dataset. This demonstrates the effectiveness of our methods. To verify the validity of generated augmentations we conduct human evaluation studies.
    \item We increase the diversity of the training dataset through augmentations and obtain comparatively better results than state-of-the-art solvers on the SVAMP challenge set.
\end{itemize}
\section{Related Work}
\textbf{Math Word Solvers:} Many research efforts have been undertaken in the recent past to solve the challenging MWP task. Broadly, these solvers can be categorized into statistical learning based and deep learning based models. Traditional approaches focused more on statistical machine learning \cite{kushman-etal-2014-learning,hosseini-etal-2014-learning} with the aim of categorizing equations into templates and extracting key patterns in the problem text. Recently, due to the advent of deep learning in NLP, solvers have witnessed a considerable increase in their performances. \cite{wang-etal-2017-deep} modelled MWP task as a sequence to sequence task and used LSTM's \cite{HochSchm97} for learning problem representations. \cite{chiang2018semantically} focused on learning representations for operators and operands.\cite{Wang2019TemplateBasedMW,xie2019goal} used tree structures for decoding process. \cite{zhang-etal-2020-graph-tree}
modelled question as a graph to map quantities and their attributes. Existing datasets which have been used as benchmark for english language includes MaWPS \cite{koncel-kedziorski-etal-2016-mawps} and Chinese language dataset Math23K \cite{wang-etal-2017-deep}. These datasets although constrained by their size deal with algebraic problems of similar difficulty levels. Recently, ASDiv \cite{miao-etal-2020-diverse} has been proposed, which has diverse problems which includes annotations for equations, problem type and grade level. Other large datasets in English language include MathQA \cite{amini2019mathqa} and Dolphin18k \cite{huang-etal-2016-well}. Although, these datasets have larger problem set but they are noisy and contain problems of varied difficulty levels.\\

\noindent\textbf{Text Data Augmentation:} To effectively train deep learning models, large datasets are required. Data augmentation is a machine learning technique that artificially enlarges the
amount of training data by means of label preserving transformations \cite{8628742}. \cite{longpre2020effective} hypothesize that textual data augmentation would only be helpful if the generated data contains new linguistic patterns that are relevant to the task and have not been seen in pre-training. In NLP, many techniques have been used for generating augmentations, \cite{wei-zou-2019-eda} introduced noise injection, deletion, insertion and swapping of words in text. \cite{Rizos2019AugmentTP} used recurrent neural networks and generative adversarial networks for short-text augmentation \cite{https://doi.org/10.48550/arxiv.2109.04775}. Recently, hard label adversarial attack models have also been used~\cite{Maheshwary2021GeneratingNL}. Other frequently used methods include inducing spelling mistakes \cite{belinkov2018synthetic}, synonym replacement \cite{zhang2016characterlevel}, identifying close embeddings from a defined search space \cite{alzantot2018generating}, round trip translations \cite{sennrich2016improving}, paraphrasing techniques \cite{kumar-etal-2019-submodular} and words predicted by language model \cite{kobayashi2018contextual} among many others. These methods are specific to the task at hand and needs to be adapted such that the generated augmentations bring diversity in the concerned dataset.
\section{Proposed Augmentation Approach}
Data augmentation generates new data by modifying existing data points through transformations based on prior knowledge about the problem domain. We introduce carefully selected transformations on well known text augmentation techniques to develop examples suited for the task of MWP. These transformations help in increasing the diversity and size of problem set in existing datasets. 
\subsection{Problem Definition}
A MWP is defined as an input of $n$ tokens, $\mathcal{P} = \{w_1,w_2..w_n\}$ where each token $w_i$ is either a numeric value or a word from a natural language. The goal is to generate a valid mathematical equation $\mathcal{E_P}$ from $\mathcal{P}$ such that the equation consists of numbers from $\mathcal{P}$, desired numerical constants and mathematical operators from the set $\{/,*,+,-,=,(,)\}$. Let $\mathcal{F}: \mathcal{P} \rightarrow \mathcal{E_P}$ be an MWP solver where $\mathcal{E_P}$ is the equation to problem $\mathcal{P}$. Our task is to generate augmented problem statement $\mathcal{P}^*$ from the original input $\mathcal{P}$ such that $\mathcal{P^*}$ is: $(1)$ semantically similar to the initial input $\mathcal{P}$, $(2)$ preserves the sequence of events in the problem statement, $(3)$ keeps the numerical values intact and $(4)$ the solution equation is same as $\mathcal{E_P}$.
\subsection{Deficiencies in Existing Models}
As showcased by \cite{patel2021nlp}, existing MWP solvers trained on benchmark datasets like MaWPS and ASDiv-A focus their attention only on certain keywords in the problem statement and do not pay much heed to the question text. We further show that even after performing significant transformations on the test set such as $(1)$ dropping the question text, $(2)$ randomly shuffling the sequence of sentences, $(3)$ random word deletion, and $(4)$ random word reordering, the solvers are still able to produce correct equations. Upon introducing these transformations we should expect a very high drop in accuracy values as the transformed problems are now distorted. Surprisingly, the decrease in accuracy scores is relatively very less than expected as shown in Table \ref{table:4}. We only observe a relatively moderate drop for word reordering. From this analysis, we can say that instead of focusing on the sequence of events, question text and semantic representation of the problem, solvers pick word patterns and keywords from the problem statement. We hypothesize that the drop in accuracy for word reordering experiment indicates that the solvers try to identify a contiguous window of words having some keywords and numbers in them, and generates equation based on these keywords. We further probe on this hypothesis by visualizing the attention weights in the experiment section.
\begin{table}[]
\small
{\renewcommand{\arraystretch}{1.1}
\begin{tabular}{@{}c|c|c|c|c@{}}
\toprule
\textbf{Dataset} & \textbf{Eval Type} & \textbf{Seq2Seq} & \textbf{GTS} & \textbf{Graph2Tree} \\ \midrule
\multirow{5}{*}{\textbf{MaWPS}} & True & 84.6 & 87.5 & 88.7 \\ \cmidrule(l){2-5} 
 & WD & \bf{80.2} & \bf{81.5} & \bf{77.3} \\ \cmidrule(l){2-5} 
 & QR & \textbf{77.4} & \textbf{82.0} & \textbf{80.2} \\ \cmidrule(l){2-5}
 & SS & \textbf{77.0} & \textbf{60.4} & \textbf{66.4} \\ \cmidrule(l){2-5}
 & WR & \textbf{54.9} & \textbf{34.8} & \textbf{39.3} \\ \midrule

\multirow{5}{*}{\textbf{ASDiv-A}} & True & 70.6 & 80.3 & 82.7 \\ \cmidrule(l){2-5} 
 & WD & \bf{60.2} & \bf{61.3} & \bf{56.7} \\ \cmidrule(l){2-5} 
 & QR & \textbf{58.7} & \textbf{52.4} & \textbf{54.1} \\ \cmidrule(l){2-5}
 & SS & \textbf{56.2} & \textbf{59.3} & \textbf{60.7} \\ \cmidrule(l){2-5}
 & WR & \textbf{47.1} & \textbf{32.3} & \textbf{34.6} \\ \bottomrule
\end{tabular}}
\caption{ Performance of solvers on modified test sets. True represents unaugmented test set. WD,QR,SS,WR represent word deletion, question reordering, sentence shuffling and word reordering respectively.}
\label{table:4}
\end{table}

\subsection{Augmentation Methods}

A MWP can also be expressed as $\mathcal{P} = (S_1,S_2..S_k,Q)$ where $Q$ is the question and $(S_1,S_2..S_k)$ are the sentences constituting the problem description. To mitigate the deficiencies in MWP solvers, we propose a two stage augmentation paradigm consisting of primary and secondary stage. In primary stage, we generate base augmentation candidates which then proceed to the secondary stage and get modified accordingly to become potential candidates. After identifying the potential candidates, we filter out the best candidates using proposed candidate selection algorithm. Table \ref{table:1} shows changes in MWP after primary and secondary stage. Following are the details:
\begin{itemize}
    \item \textbf{Primary Stage:} In the primary stage, our focus is on inducing variations in the question text $Q$ of a given problem statement $\mathcal{P}$. For this, we first generate $n$ base candidates ${\{b_1,b_2,...,b_n\}}$ from $Q$ using $T5$ paraphrasing model \cite{raffel2020exploring}.
    The key intuition behind this step is to ensure that each augmentation of a given problem has a different question text. This will empower the solver to learn variations from the question text as well.
    \item \textbf {Secondary Stage:} After the generation of base candidates, we implement augmentation methods to generate potential candidates. These methods although well known, require careful tuning to adapt for MWP generation. Table \ref{table:3} showcases MWP examples and their generated augmentations. Detailed description of these techniques follow.
\end{itemize}

\begin{table*}[]
\small
\begin{tabular}{@{}p{2cm} p{4cm} p{9cm}@{}}
\toprule
\textbf{Category} & \textbf{Augmentation Method} & \textbf{Example} \\ \midrule
\multirow{4}{*}{\textbf{\begin{tabular}[c]{@{}l@{}}Paraphrasing\\ Methods\end{tabular}}} & \multirow{2}{*}{\textbf{\textcolor{red}{Round trip Translation}}} & \textbf{Original:} The schools debate team had 4 boys and 6 girls on it. If they were split into groups of 2, how many groups could they make ? \\
 &  & \textbf{Augmented:} The school \textcolor{violet}{discussion group consisted} of 4 boys and 6 girls. If they are \textcolor{violet}{divided} into groups of 2 . \textcolor{violet}{How many groups could they have created ?} \\ \cmidrule(l){2-3} 
 & \multirow{2}{*}{\textbf{\textcolor{red}{Problem Reordering}}} & \textbf{Original:} Lucy has an aquarium with 5 fish . She wants to buy 1 more fish . How many fish would Lucy have then ?
\\
 &  & \textbf{Augmented:} \textcolor{violet}{If} lucy has an aquarium with 5 fish \textcolor{violet}{and} she wants to buy 1 more fish \textcolor{violet}{then} \textcolor{violet}{how many fish would lucy have ?}
 \\ \hline 
\multirow{6}{*}{\textbf{\begin{tabular}[c]{@{}l@{}}Substitution\\ Methods\end{tabular}}} & \multirow{2}{*}{\textbf{\textcolor{red}{Fill Masking}}} & \textbf{Original:} There are 8 walnut trees currently in the park . Park workers will plant 3 more walnut trees today . How many walnut trees will the park have when the workers are finished ?
\\
 &  & \textbf{Augmented:} There are 8 walnut trees currently in the park . Park workers will plant 3 more walnut trees \textcolor{violet}{soon} . How many walnut trees will the park have \textcolor{violet} {after the workers are finished ?}
 \\ \cmidrule(l){2-3} 
 & \multirow{2}{*}{\textbf{\textcolor{red}{Named-Entity Replacement}}} & \textbf{Original:} Sally found 7 seashells , Tom found 12 seashells , and Jessica found 5 seashells on the beach . How many seashells did they find together ?\\
 &  & \textbf{Augmented:} \textcolor{violet}{Edd} found 7 seashells , \textcolor{violet}{Alan} found 12 seashells , and \textcolor{violet}{Royal} found 5 seashells on the beach . \textcolor{violet}{How many seashells were found together ?}
 \\ \cmidrule(l){2-3} 
 & \multirow{2}{*}{\textbf{\textcolor{red}{Synonym Replacement}}} & \textbf{Original:} Katie 's team won their dodgeball game and scored 25 points total . If Katie scored 13 of the points and everyone else scored 4 points each , how many players were on her team ?
 \\
 &  & \textbf{Augmented:} Katie's \textcolor{violet}{group} won their \textcolor{violet}{rumble} game and scored 25 points total . If Katie scored 13 of the points and \textcolor{violet}{all} else scored 4 points each, How many players was on her \textcolor{violet}{group} ?
\\
 \bottomrule
\end{tabular}
\caption{Augmentation examples from all proposed methods. Coloured text represents the changes in problem statement.}
\label{table:3}
\end{table*}
   
\subsubsection{Paraphrasing Methods}
Paraphrasing has proved to be an effective way of generating text augmentations \cite{witteveen-andrews-2019-paraphrasing}.
It generates samples having diverse sentence structures and word choices while preserving the semantic meaning of the text. These additional samples guide the model to pay attention to not only the keywords but its surroundings as well. This is particularly beneficial for the task of MWP solving, where most of the problem statements follow a general structure.\\

\textbf{Problem Reordering:} Given original problem statement $\mathcal{P} = (S_1,S_2,...S_k,Q)$, we alter the order of problem statement such that $\mathcal{P^*} = (Q,S_1,S_2,...,S_k)$. To preserve the semantic and syntactic meaning of problem statement we use filler phrases like 'Given that' and 'If-then'. To make these paraphrases more fluent, we use named entity recognition and co-reference resolution to replace the occurrences of pronouns with their corresponding references. Please note that this method is better than random shuffling of sentences as it preserves the sequence of events in the problem statement.\\

\textbf{Round Trip Translations:} Round trip translations, more commonly referred as back-translation is an interesting method to generate paraphrases. This idea has evolved as a result of the success of machine translation models  \cite{Wu2016GooglesNM}. In this technique, sentences are translated from their original language to foreign languages and then translated back to the original language. This round trip can be between multiple languages as well. The motivation behind using this technique is to utilize the different structural constructs and linguistic variations present in other languages.\\
Back-translation is known to diverge uncontrollably \cite{tan2019learning} for multiple round trips. This may lead to change in the semantics of the problem statement. Numerical quantities are fragile to translations and their order and representation may change. To overcome these challenges, we worked with languages that have structural constructs similar with English. For instance, languages like Finnish which are gender neutral, can become problematic as they can lead to semantic variance in augmented examples. To preserve numerical quantities, we replace them with special symbols and keep a map to restore numerical quantities in the generated paraphrases.
We have used the following round trips:

\noindent \textit{English - Russian - English:} Although Russian is linguistically different from English, we still chose it as word order does not affect the syntactic structure of a sentence in Russian language \cite{voita2019contextaware}. For single round trip, we preferred Russian as it has the potential to generate different paraphrase structures.

\noindent \textit{English - German - French - English:} German and french are structurally similar to English language \cite{kim-etal-2019-pivot}, we chose them for multiple round trips to both maintain semantic in-variance and induce minor alterations in the paraphrases.

\subsubsection{Substitution Methods}
In this class of methods, the focus is on generating  variations of the problem statement by identifying and substituting some of the keywords such that the augmentations are semantically and syntactically correct, with the equation labels preserved. Substitution is effective for MWP solving as it guides the solvers focus away from certain keywords, allowing it to distribute its attention and generalize better. We propose the following methods:\\

\textbf{Fill-Masking:}
In this technique, we model the challenge of generating candidates as a masked language modelling problem. Instead of randomly choosing words for masking, we use part of speech tags to focus on nouns and adjectives, preferably in the vicinity of numerical quantities. We replace these identified keywords with mask tokens. These masked candidate sentences are then passed through a masked language model \cite{Devlin2019BERTPO} and suitable words are filled in masked positions to generate our candidate sentences.\\

\textbf{Synonym Replacement:}
In this method, after stop-word removal, we select keywords randomly for substitution. Unlike fill-mask technique, where masked language models were deployed, here we use Glove embeddings \cite{pennington2014glove} to find the top $k$ candidates that are close synonyms of the keywords. To ensure syntactic correctness in candidates, we maintain the part of speech tags for the substitute candidates. These synonyms are then used to substitute the keywords in the problem  statement and generate augmented candidates.\\

\textbf{Named-Entity Replacement:} A common occurrence in MWP is the usage of named entities. These entities play a crucial role in stating the problem statement, but the solution equations do not change on altering these entities. Following this insight, we first identify the named entities\footnote{https://www.nltk.org/} such as person, place and organizations present in the problem statement. Then we replace these named entities with their corresponding substitutes, like a person's name is replaced by another person's name to generate the potential candidates.\\
Table \ref{table:2} reports the statistics of augmented datasets on both MaWPS and ASDiv-A.
All the techniques described in paraphrasing and substitution methods are used for generating the potential candidates for a problem statement. After generation of the potential candidates for augmenting a problem statement, the best possible candidate is selected by using Algorithm \ref{algorithm:1}. Key motivation behind developing this algorithm is to select candidates on which the solver does not perform well and which are similar to the original problem statement.\\

We use negative log likelihood as the loss function $\mathcal{L}$ and Sentence-BERT  \cite{reimers2019sentencebert} fine tuned on MWP equation generation task as sentence embedding generator $\mathcal{S}$. We calculate the similarity of each candidate embedding with the original problem representation using cosine similarity as shown in Line $3$ of the algorithm. Further, for each candidate sentence, we evaluate their loss values and select the candidate with the maximum mean normalized loss and similarity score.
\begin{table}[ht]
\small
\centering
{\renewcommand{\arraystretch}{1.1}
\begin{tabular}{|c|c | c|}
\hline
\textbf{Dataset} & \textbf{Problem Size} & \textbf{Vocabulary Size}\\ [0.5ex]
\hline
MaWPS & 2,373 & 2,632\\
ASDiv-A & 1,213 & 2,893\\
Paraphrase & 5,909 & 3,832\\
Substitution & 6,647 & 3,923\\
Combined-MaWPS & \bf{10,634} & \bf{5,626}\\
Combined-ASDiv & \bf{5,312} & \bf{6,109}\\
\hline
\end{tabular}}
\caption{Statistics of augmented dataset compared with MaWPS and ASDiv-A. Combined-Dataset represents combination of Paraphrase and Substitution methods.}
\label{table:2}
\end{table}
\begin{algorithm}[h!]
\caption{MWP Candidate Selection Algorithm}
\textbf{Requires: }$\mathcal{M}$ is augmentation method,~$\mathcal{S}$ is similarity model,~$\mathcal{F}$ is solver model,~$\mathcal{L}$ is Loss function.\\
\textbf{Input: }Problem text $\mathcal{P}$\\
\textbf{Output: }Augmented Text $\mathcal{P^*}$
\begin{algorithmic}[1]
\State{$\mathcal{E}_{P} \gets \mathcal{F}(\mathcal{P})$}
\State{$Candidates \gets \mathcal{M}\mathcal{(P)}$}
\For{$C_j\ in\ Candidates :$}
% \STATE {$\mathcal{E}_j \gets \mathcal{F}(C_j)$}
\State{$S_j \gets \mathcal{S}(C_j,\mathcal{P})$}
\State{$L_{j} \gets (\mathcal{L}(C_j) - \mathcal{L}(P))/\mathcal{L}(P)$}
\State{$CandidateScore.add(S_j*L_j)$}
\EndFor
\State{$\mathcal{P^*} = \underset{C_j}{\arg\max}\ CandidateScore(C_j)$}
\State{\textbf{end}}
\end{algorithmic}
\label{algorithm:1}
\end{algorithm}
\section{Experiments}
\textbf{Datasets and Models:} To showcase the effectiveness of proposed augmentation methods, we select three state-of-the-art MWP solvers: $(1)$ \emph{Seq2Seq}~\cite{wang-etal-2017-deep} having an LSTM encoder and an attention based decoder. $(2)$ \emph{GTS}~\cite{xie2019goal} having an LSTM encoder and a tree based decoder and $(3)$ \emph{Graph2tree}~\cite{zhang-etal-2020-graph-tree} consists of a both tree based encoder and decoder. \emph{Seq2Seq} serves as our base model for experimentation. Many existing datasets are not suitable for our analysis as either they are in Chinese~\citep{wang-etal-2017-deep} or they have problems of higher complexities ~\citep{huang-etal-2016-well} . We conduct experiments across the two largest available English language datasets satisfying our requirements: $(1)$ \emph{MaWPS}~\cite{koncel-kedziorski-etal-2016-mawps} containing $2,373$ problems $(2)$ \emph{ASDiv-A}~\cite{miao-etal-2020-diverse} containing $1,213$ problems. Both datasets have MWPs with linear equation in one variable.\\

\noindent\textbf{Experiment Setup:} We train and evaluate the three solvers on both MaWPS and ASDiv-A using five fold cross validation. Evaluation is conducted on both original and augmented datasets. We use the same hyperparameter values as recommended in the original implementation of these solvers. Further, each solver has been trained from scratch and by using BERT embeddings~\cite{devlin2019bert}. We also evaluate the models on \emph{SVAMP}~\cite{patel2021nlp} challenge set. This test set has been designed specifically to examine the robustness and adaptability of the solvers. Ablation studies have been conducted to assess the effectiveness of candidate selection algorithm and augmentation techniques.
\begin{table}[]
\small
{\renewcommand{\arraystretch}{0.8}
\begin{tabular}{@{}c|c|c|c|c@{}}
\toprule
\textbf{Dataset} & \textbf{Evaluation Type} & \textbf{Seq2Seq} & \textbf{GTS} & \textbf{G2T} \\ \midrule
\multirow{4}{*}{\textbf{MaWPS}} & True & 84.6 & 87.5 & 88.7 \\ \cmidrule(l){2-5} 
 & Paraphrasing & \bf{88.3} & \bf{90.4} & \bf{92.6} \\ \cmidrule(l){2-5} 
 & Substitution & \textbf{89.2} & \textbf{89.7} & \textbf{91.7} \\ \cmidrule(l){2-5}
 & Combined & \textbf{91.3} & \textbf{92.6} & \textbf{93.5} \\ \midrule
\multirow{4}{*}{\textbf{ASDiv-A}} & True & 70.6 & 80.3 & 82.7 \\ \cmidrule(l){2-5} 
 & Paraphrasing & \bf{75.6} & \bf{84.2} & \bf{83.6} \\ \cmidrule(l){2-5} 
 & Substitution & \textbf{73.2} & \textbf{83.3} & \textbf{84.1} \\ \cmidrule(l){2-5}
 & Combined & \textbf{78.2} & \textbf{85.9} & \textbf{86.3} \\ \bottomrule
\end{tabular}}
\caption{Result of augmentation methods. True is original dataset, Combined is combination of paraphrasing and substitution. G2T represents Graph2Tree solver.}
\label{table:8}
\end{table}
\subsection{Results and Analysis}
Table \ref{table:8} shows the result of proposed methods. These results have been reported on BERT embeddings. Table \ref{table:11} shows a comparison between training from scratch and using BERT embeddings. By training these state-of-the-art models on the augmented dataset we achieve better results for both MaWPS and ASDiv-A. On average, we were able to increase the accuracy significantly by more than five percentage points. Both paraphrasing and substitution methods have performed well independently and in combination. Further, we conduct ablation studies to analyze the performance of each augmentation method.
\begin{table}[]
\centering
\small
{\renewcommand{\arraystretch}{1}
\begin{tabular}{@{}l@{}}
\toprule
\begin{tabular}[c]{@{}l@{}}\textcolor{red}{Problem 1:} Ricardo was making baggies of cookies with\\ 5 cookies in each bag. If he had 7 chocolate chip cookies\\ and 3 oatmeal cookies, how many baggies could he make ?\\
\textcolor{violet}{Solution Equation:} X = (7+3)/5\\
\textcolor{orange}{Pre Augmentation Equation:} X = (7/3)/3\\
\textcolor{blue}{Post Augmentation Equation:} X = (7+3)/5\\
\end{tabular} \\ \midrule
\begin{tabular}[c]{@{}l@{}}\textcolor{red}{Problem 2:} For halloween Destiny bought 9 pieces of candy.\\ She ate 3 pieces the first night and then her sister\\ gave her 2 more pieces. How many pieces of candy does\\ Destiny have now ?\\
\textcolor{violet}{Solution Equation:} X = 9-3+2\\
\textcolor{orange}{Pre Augmentation Equation:} X = ((9+3-3\\
\textcolor{blue}{Post Augmentation Equation:} X = (9+3-2) \\\end{tabular} \\ \midrule
\begin{tabular}[c]{@{}l@{}}\textcolor{red}{Problem 3 :} Audrey needs 6 cartons of berries to make\\ a berry cobbler. She already has 2 cartons of strawberries\\ and 3 cartons of blueberries. How many more cartons of\\ berries should Audrey buy ?\\
\textcolor{violet}{Solution Equation:} X = 6-2-3\\
\textcolor{orange}{Pre Augmentation Equation:} X = (6-(2)+3)\\
\textcolor{blue}{Post Augmentation Equation:} X = 6-(2+3) \\\end{tabular} \\
\bottomrule
\end{tabular}}
\caption{Examples illustrating equation results before and after training on the full augmented dataset.}
\label{table:9}
\end{table}
In Table \ref{table:9} we illustrate some examples on which existing models generate incorrect equations. However, after being trained with augmented dataset they generate correct equations. Additionally, in Problem 2 the base model generates syntactically incorrect solution, but post augmentation it generates syntactically correct equation. These examples show the increased robustness and solving abilities of solvers.
\begin{table}[]
\centering
\small
{\renewcommand{\arraystretch}{1}
\begin{tabular}{@{}l@{}}
\toprule
\begin{tabular}[c]{@{}l@{}} \textcolor{red}{Problem:} Gavin has 6 shirts . 3 are \colorbox{brown}{blue} the \colorbox{pink}{rest} \colorbox{orange}{are}\\ green. How many green shirts does Gavin have ?\\
\textcolor{red}{Mean attention values:} \colorbox{brown}{0.27}  \colorbox{orange}{0.14}  \colorbox{pink}{0.08}\\\\
\textcolor{red}{Problem:}  \colorbox{brown}{Gavin} has 6 shirts . 3 are blue the \colorbox{pink}{rest} are\\ green. How \colorbox{orange}{many} green shirts does Gavin have ?\\
\textcolor{red}{Augmented mean attention values :} \colorbox{brown}{0.23}  \colorbox{orange}{0.18}  \colorbox{pink}{0.11}  
\end{tabular} \\ \midrule
\begin{tabular}[c]{@{}l@{}} 
\textcolor{red}{Problem:} There are 3 pencils \colorbox{orange}{in} the \colorbox{brown}{drawer}. Sara\\ placed 7\colorbox{pink}{more} pencils in the drawer. How many pencils\\ are there in all ?\\
\textcolor{red}{Mean attention values:} \colorbox{brown}{0.45}  \colorbox{orange}{0.11}  \colorbox{pink}{0.05}\\
\\
\textcolor{red}{Problem:} There are 3 \colorbox{pink}{pencils} in the drawer. Sara\\ \colorbox{brown}{placed} 7 more pencils in the drawer. How many pencils\\ are there in \colorbox{orange}{all} ?\\
\textcolor{red}{Augmented mean attention values :} \colorbox{brown}{0.31}  \colorbox{orange}{0.16}  \colorbox{pink}{0.09} \\ \end{tabular} \\
\bottomrule
\end{tabular}}

\caption{Examples illustrating distribution of top three attention weights before and after training on the full augmented dataset.}
\label{table:5}
\end{table}
\\

\noindent\textbf{Attention Visualizations:} 
Through this investigation, we aim to ascertain our hypothesis that to generate equations MWP solvers focus only on certain keywords and patterns in a region. They ignore essential information like semantics, sequence of events and content of the question text present in the problem statement. In Table \ref{table:5}, we show some sample problem statements with their attention weights. These weights are generated during the decoding process using Luong attention mechanism \cite{luong2015effective}. Moreover, to illustrate the effectiveness of our augmentation techniques, we show the distribution of attention weights for models trained on the augmented dataset. We can infer from the examples showcased in Table \ref{table:5} that before augmentation the focus of the solver is limited to a fixed region around numerical quantities and it does not pay heed to the question text. However, after training on the augmented dataset the solver has a better distribution of attention weights, the weights are not localised and and the model is also able to pay attention on the question text. \\

\noindent\textbf{Ablation Studies:}
To assert the effectiveness of our methods, we conduct the following ablations:\\
\textit{Candidate Selection Algorithm:} For testing the usefulness of candidate selection algorithm, we compare it with a random selection algorithm. In this, we randomly select one of the possible candidates as augmented problem statement. We evaluate the accuracy of models trained on the augmented datasets, generated using both the algorithms.
Result in Table \ref{table:6}, shows that candidate selection algorithm performs better than random selection algorithm and this demonstrates the effectiveness of our algorithm.\\

\textit{Augmentation Methods:} To examine the effectiveness of proposed augmentation techniques, we evaluate the models on each of the proposed techniques independently and report the results in Table \ref{table:7}. Although, all the methods contribute towards increase in accuracy but Round trip translations and synonym replacement perform marginally better than others. This behaviour can be linked to the structural diversity and keyword sensitivity that round trip translations and synonym replacement bring respectively~\cite{feng2021survey}.

\begin{table}[]
\small
{\renewcommand{\arraystretch}{0.9}
\begin{tabular}{@{}c|c|c|c|c@{}}
\toprule
\textbf{Method} & \textbf{Eval Type} & \textbf{Seq2Seq} & \textbf{GTS} & \textbf{Graph2Tree} \\ \midrule
\multirow{4}{*}{\textbf{RSA}} & True & 84.6 & 87.5 & 88.7 \\ \cmidrule(l){2-5} 
 & Paraphrasing & \bf{85.3} & \bf{88.1} & \bf{89.2} \\ \cmidrule(l){2-5} 
 & Substitution & \textbf{86.8} & \textbf{87.3} & \textbf{87.9} \\ \cmidrule(l){2-5}
 & Combined & \textbf{87.0} & \textbf{89.2} & \textbf{89.5} \\ \midrule

\multirow{4}{*}{\textbf{CSA}} & True & 84.6 & 87.5 & 88.7 \\ \cmidrule(l){2-5} 
 & Paraphrasing & \bf{88.3} & \bf{90.4} & \bf{92.6} \\ \cmidrule(l){2-5} 
 & Substitution & \textbf{89.2} & \textbf{89.7} & \textbf{91.7} \\ \cmidrule(l){2-5}
 & Combined & \textbf{91.3} & \textbf{92.6} & \textbf{93.5} \\ \bottomrule
\end{tabular}}
\caption{Ablation Study for Random Selection Algorithm (RSA) and Candidate Selection Algorithm (CSA).}
\label{table:6}
\end{table}
\begin{table}[ht]
\centering
\small
{\renewcommand{\arraystretch}{1.1}
\begin{tabular}{|c|c | c| c|}
\hline
\textbf{Augmentation } & \textbf{Seq2Seq} & \textbf{GTS} & \textbf{Graph2Tree}\\
\hline
True & 84.6 & 87.5 & 88.7\\
RRT & \bf{86.5} & 89.1 & \bf{91.6} \\
PR & 85.9 & 88.4 & 90.7 \\
FM & 84.8 & 87.2 & 89.1 \\
SR & 85.2 & \bf{90.1} & 91.2\\
NER & 86.1 & 88.3 & 89.7\\
\hline
\end{tabular}}
\caption{Result of Ablation study for each augmentation method. True represents unaugmented MaWPS dataset, RRT, PR, FM, SR, NER represents round trip translations, problem reordering, fill masking,synonym replacement and named entity replacement respectively.}
\label{table:7}
\end{table}
\begin{table}[ht]
\centering
\small
{\renewcommand{\arraystretch}{1.1}
\begin{tabular}{|c|c | c| c|}
\hline
\textbf{Augmentation } & \textbf{Seq2Seq} & \textbf{GTS} & \textbf{Graph2Tree}\\
\hline
True & 37.5 & 39.6 & 41.2\\
MaWPS(P+S) & 39.2 & 40.1 & 42.3 \\
ASDiv-A(P+S) & 37.8 & 40.4 & 42.1\\
Combined & \bf{40.2} & \bf{41.3} & \bf{43.8}\\
\hline
\end{tabular}}
\caption{Result of augmentations on SVAMP Challenge Set. P and S represent paraphrasing and substitution methods. Combined represents augmented MaWPS and ASDiv-A. True is combined MaWPS and ASDiv-A.}
\label{table:10}
\end{table}
\noindent\textbf{SVAMP Challenge Set:} SVAMP \cite{patel2021nlp} is a manually curated challenge test set consisting of $1,000$ math word problems. These problems have been cherry picked from MaWPS and ASDiv-A, then altered manually to modify the semantics of question text and generate additional equation templates. This challenge set is suitable for evaluating a solver's performance as it modifies problem statements such that solver's generalization can be checked. The results are shown in Table \ref{table:10}.
Although, our proposed augmented dataset has very limited equation templates, still it performs comparatively better than state-of-the-art models on SVAMP challenge set. This result signifies the need for a larger and diverse dataset with enhanced variety of problems. Further, it demonstrates the effectiveness of our method which is able to perform better on SVAMP test set and increase model's accuracy despite the challenges.\\
\begin{table}[ht]
\centering
\small
\begin{tabular}{|c|c | c | c | c|}
\hline
\multirow{2}{*}{\begin{tabular}{c}\textbf{Augmentation}\\\textbf{Method}\end{tabular}} & \multicolumn{2}{c|}{\textbf{MaWPS}} & \multicolumn{2}{c|}{\textbf{ASDiv-A}}\\ \cline{2-5} 
& Scratch & BERT & Scratch & BERT \\
\hline
True & 77.2  & \bf{84.6} & 53.2 & \bf{70.6} \\
Paraphrasing & 79.8 & \bf{88.3} & 58.1 & \bf{75.6}\\
Substitution & 81.3 & \bf{89.2} & 57.3 & \bf{73.2}\\
Combined & 82.7 & \bf{91.3} & 60.4 & \bf{78.2}\\
\hline
\end{tabular}
\caption{Performance comparison of baseline model trained from scratch and trained using BERT embeddings. True represents unaugmented dataset.}
\label{table:11}
\end{table}\\

\noindent \textbf{BERT Embeddings:} We train the solvers in two different settings, using pre-trained BERT embeddings and training from scratch. We chose BERT specifically as we require contextual embeddings which could be easily adapted for the task of MWP. Moreover, existing models have also shown results using BERT and it would be fair to compare their performances when trained using similar embeddings. Results obtained are shown in Table \ref{table:11}. We observe that for solver's trained using BERT, accuracy is higher than models trained from scratch.\\

\noindent\textbf{Human Evaluation:}
To verify the quality of augmented examples, we conduct human evaluation. The focus of this evaluation is: $(1)$ To check if the augmentations will result in the same linear equation as present in the original problem statement, $(2)$ To evaluate if the numerical values for each augmentation example is preserved, $(3)$ Evaluate each sample in the range $0$ to $1$ for its semantic similarity with the original problem statement, $(4)$ On a scale of $1$ to $5$ rate each augmented example for its grammatical correctness. We conduct the human evaluations on randomly shuffled subsets consisting of around $40\%$ of the total augmented examples for both the datasets. This process is repeated three times with different subsets, five human evaluators evaluate each example in all subsets, and the mean results are computed as shown in Table \ref{table:12}.
\begin{table}[ht]
\centering
\small
\begin{tabular}{|c|c | c | c | c|}
\hline
\multirow{2}{*}{\begin{tabular}{c}\textbf{Evaluation}\\\textbf{Criteria}\end{tabular}} & \multicolumn{2}{c|}{\textbf{MaWPS}} & \multicolumn{2}{c|}{\textbf{ASDiv-A}}\\ \cline{2-5} 
& Para & Sub & Para & Sub \\
\hline
Preserves Equation & 92.3\% & 89.5\% & 93.6\% & 90.1\%\\
Preserves Numbers & 88.4\% & 91.2\% & 87.3\% & 90.3\%\\
Semantic Similarity & 0.96 & 0.89 & 0.91 & 0.87\\
Syntactic Similarity & 4.67 & 4.36 & 4.59 & 4.33\\
\hline
\end{tabular}
\caption{Human Evaluation scores on augmentated dataset. Para and Sub represents paraphrasing and substitution methods respectively.}
\label{table:12}
\end{table}
\section{Future Work and Conclusion}
 We showcase that the existing MWP solvers are not robust and do not generalize well on even simple variations of the problem statement. In this work, we have introduced data augmentation techniques for generation of diverse math word problems. We were able to enhance the size of existing dataset by 5 folds and significantly increase the performance of state-of-the-art solvers by over 5 percentage points. Future works could focus on developing techniques to generate data artificially and making robust MWP solvers.
 \section{Acknowledgment}
 We thank IHub-Data, IIIT Hyderabad\footnote{\href{https://ihub-data.iiit.ac.in/}{https://ihub-data.iiit.ac.in/}} for financial support. 
\bibliography{anthology,custom}
\bibliographystyle{acl_natbib}
\newpage
\section{Appendix}
\subsection{Implementation Details}
For conducting our experiments we have used two Boston SYS-7048GR-TR nodes equipped with NVIDIA GeForce GTX 1080 Ti computational GPU’s having 11GB of GDDR5X RAM. All implementations of training and testing is coded in Python with Pytorch framework.  The number of parameters range from 20M to 130M for different models. We use negative log likelihood as the loss criterion. Hyper-parameter values were not modified, and we follow the recommendations of the respective models. To reduce carbon footprint from our experiments, we run the models only on a single fold for searching hyperparameter values. 
We chose the number of base candidates after primary stage $n$ as 7. Generating augmentation examples using Paraphrasing Methods took around 12 minutes on average for MaWPS and 8 minutes for ASDiv-A datasets. Substitution methods took around 5 minutes on average for both MaWPS and ASDiv-A dataset. The experiments conducted by us are not computation heavy. Each of the state-of-the-art models get trained within 5 hrs of time, with Graph2Tree taking the maximum time.
\subsection{Additional Augmented Examples}
In this section, we present some additional valid as well as invalid augmented examples. Additionally, we also show some more examples with their attention weights. Table \ref{table:a1} shows some additional examples with their attention weight distribution. These weights have been shown for the base model trained before augmentation and after augmentation on MaWPS dataset.
\noindent Table \ref{table:a2} illustrates some additional problem statements for all the techniques in paraphrasing methods and substitution methods. In Table \ref{table:a3},
we present some invalid augmented examples which do not satisfy our human evaluation criteria. These examples are such that they alter the semantics of the original problem statement.
\begin{table}[]
\centering
\small
{\renewcommand{\arraystretch}{1}
\begin{tabular}{@{}l@{}}
\hline
\begin{tabular}[c]{@{}l@{}} \textcolor{red}{Problem:} A magician \colorbox{orange}{was} selling magic card \colorbox{brown}{decks} for 2\\ dollars \colorbox{pink}{each}. If he started with 25 decks and by the end of\\ the day he had 4 left, how much money did he earn ?\\
\textcolor{red}{Mean attention values:} \colorbox{brown}{0.34}  \colorbox{orange}{0.11}  \colorbox{pink}{0.09}\\\\
\textcolor{red}{Problem:} A magician was selling magic \colorbox{brown}{card} decks for 2\\ \colorbox{pink}{dollars} each. If he started with 25 decks and by the end of\\ the day he had 4 left,  how \colorbox{orange}{much} money did he earn ?\\
\textcolor{red}{Augmented mean attention values :} \colorbox{brown}{0.19}  \colorbox{orange}{0.18}  \colorbox{pink}{0.15}  
\end{tabular} \\ \midrule
\begin{tabular}[c]{@{}l@{}} 
\textcolor{red}{Problem:} There \colorbox{pink}{are} 18 pencils in the drawer and 6 pencils\\ \colorbox{orange}{on} the desk. Dan placed 4 \colorbox{brown}{pencils} on the desk. How many\\ pencils are there in total ?\\

\textcolor{red}{Mean attention values:} \colorbox{brown}{0.21}  \colorbox{orange}{0.16}  \colorbox{pink}{0.06}\\
\\
\textcolor{red}{Problem:} There are 18 pencils \colorbox{pink}{in} the drawer and 6 pencils\\ on the desk.\\\colorbox{brown}{Dan} placed 4 pencils on the desk. How many\\ pencils are there in \colorbox{orange}{total} ?\\

\textcolor{red}{Augmented mean attention values :} \colorbox{brown}{0.29}  \colorbox{orange}{0.19}  \colorbox{pink}{0.12} \\ \end{tabular} \\
\midrule
\begin{tabular}[c]{@{}l@{}} \textcolor{red}{Problem:} Dan has 12 violet marbles, he \colorbox{orange}{gave} Mary 4\\ of the \colorbox{brown}{marbles}. How many violet marbles does \colorbox{pink}{he}\\ now have ?
 \\
\textcolor{red}{Mean attention values:} \colorbox{brown}{0.23}  \colorbox{orange}{0.21}  \colorbox{pink}{0.17}\\\\
\textcolor{red}{Problem:} Dan has 12 \colorbox{orange}{violet} marbles , he \colorbox{brown}{gave} Mary 4\\ of the marbles. How \colorbox{pink}{many} violet marbles does he\\ now have ?
\\
\textcolor{red}{Augmented mean attention values :} \colorbox{brown}{0.23}  \colorbox{orange}{0.18}  \colorbox{pink}{0.11}  
\end{tabular} \\ \midrule
\begin{tabular}[c]{@{}l@{}} 
\textcolor{red}{Problem:} Angela has 7 tickets . \colorbox{pink}{Annie} gives Angela\\ 5 \colorbox{brown}{more}. How many \colorbox{orange}{tickets} does Angela have in all ?\\
\textcolor{red}{Mean attention values:} \colorbox{brown}{0.30}  \colorbox{orange}{0.19}  \colorbox{pink}{0.15}\\
\textcolor{red}{Problem:} Angela has 7 tickets . Annie \colorbox{pink}{gives} Angela\\ 5 \colorbox{orange}{more}. How many tickets does Angela have in \colorbox{brown}{all} ?\\
\textcolor{red}{Augmented mean attention values :} \colorbox{brown}{0.29}  \colorbox{orange}{0.21}  \colorbox{pink}{0.14} \\ \end{tabular} \\
\midrule
\begin{tabular}[c]{@{}l@{}} \textcolor{red}{Problem:} Maria \colorbox{pink}{had} 5 \colorbox{orange}{bottles} of water \colorbox{brown}{in} her fridge.\\ If she drank 1 of them and then bought 2 more, how many\\ bottles would she have ?\\
\textcolor{red}{Mean attention values:} \colorbox{brown}{0.48}  \colorbox{orange}{0.14}  \colorbox{pink}{0.04}\\
\textcolor{red}{Problem} Maria had 5 bottles of water in her fridge.\\ If she drank 1 of them and \colorbox{pink}{then} bought 2 \colorbox{brown}{more} , how\\ \colorbox{orange}{many} bottles would she have ?\\
\textcolor{red}{Augmented mean attention values :} \colorbox{brown}{0.23}  \colorbox{orange}{0.17}  \colorbox{pink}{0.11}  
\end{tabular} \\ 
\bottomrule
\end{tabular}}
\caption{Examples illustrating distribution of top three attention weights before and after training on the full augmented dataset.}
\label{table:a1}
\end{table}
\begin{table*}[]
\small
\begin{tabular}{@{}p{2cm} p{4cm} p{9cm}@{}}
\toprule
\textbf{Category} & \textbf{Augmentation Method} & \textbf{Example} \\ \midrule
\multirow{4}{*}{\textbf{\begin{tabular}[c]{@{}l@{}}Paraphrasing\\ Methods\end{tabular}}} & \multirow{2}{*}{\textbf{\textcolor{red}{Round trip Translation}}} & \textbf{Original:} Alyssa's dog had puppies. She gave 2 to her friends. She now has 3 puppies. How many puppies did she have to start with ?
 \\
 &  & \textbf{Augmented:} Alyssa's dog had puppies . \textcolor{violet}{She gave her friends 2} . She now has 3 puppies . \textcolor{violet}{How many puppies did she start ?}
 \\ \cmidrule(l){2-3} 
 & \multirow{2}{*}{\textbf{\textcolor{red}{Problem Reordering}}} & \textbf{Original:} Rachel was organizing her book case making sure each of the shelves had exactly 3 books on it. If she had 4 shelves of mystery books and 2 shelves of picture books , how many books did she have total ?
\\
 &  & \textbf{Augmented:}\textcolor{violet}{How many books did she have given that} rachel was organizing her book case making sure each of the shelves had exactly 3 books on it and she had 4 shelves of mystery books and 2 shelves of picture books .
 \\ \hline 
\multirow{6}{*}{\textbf{\begin{tabular}[c]{@{}l@{}}Substitution\\ Methods\end{tabular}}} & \multirow{2}{*}{\textbf{\textcolor{red}{Fill Masking}}} & \textbf{Original:} A cell phone company has a total of 1000 customers across the world . If 740 of its customers live in the United States , how many of its customers live in other countries ?
\\
 &  & \textbf{Augmented:} A \textcolor{violet}{mobile} phone \textcolor{violet}{firm} has a network of 1000 customers across the world . If 740 of its customers live in the \textcolor{violet}{US}, \textcolor{violet}{How many customers live in other locations?}\\
\cmidrule(l){2-3} 
 & \multirow{2}{*}{\textbf{\textcolor{red}{Named-Entity Replacement}}} & \textbf{Original:} Daniel had some noodles. He gave 20 noodles to William. Now Daniel only has 11 noodles. How many noodles did Daniel have to begin with ?
\\
 &  & \textbf{Augmented:} Matt had some noodles. He gave 20 noodles to Zeal. Now Matt only has 11 noodles. How many noodles did Matt have initially ? \textcolor{violet}{Edd} found 7 seashells , \textcolor{violet}{Alan} found 12 seashells , and \textcolor{violet}{Royal} found 5 seashells on the beach . \textcolor{violet}{How many seashells were found together ?}
 \\ \cmidrule(l){2-3} 
 & \multirow{2}{*}{\textbf{\textcolor{red}{Synonym Replacement}}} & \textbf{Original:} There are 5 rulers in the drawer. Tim took 3 rulers from the drawer. How many rulers are now in the drawer ?
 \\
 &  & \textbf{Augmented:} There are 5 \textcolor{violet}{consonants} in the drawer. Tim went 3 \textcolor{violet}{consonants} from the drawer. \textcolor{violet}{How many other consonants are in the drawer now ?}
\\
 \bottomrule
\end{tabular}
\caption{Valid Augmentation examples from all proposed methods. Coloured text represents the changes in problem statement.}
\label{table:a2}
\end{table*}

\begin{table*}[]
\small
\begin{tabular}{@{}p{2cm} p{4cm} p{9.3cm}@{}}
\toprule
\textbf{Category} & \textbf{Augmentation Method} & \textbf{Example} \\ \midrule
\multirow{4}{*}{\textbf{\begin{tabular}[c]{@{}l@{}}Paraphrasing\\ Methods\end{tabular}}} & \multirow{2}{*}{\textbf{\textcolor{red}{Round trip Translation}}} & \textbf{Original:} Kimberly went to the store 6 times last month . She buys 9 peanuts each time she goes to the store . How many peanuts did Kimberly buy last month ?
 \\
 &  & \textbf{Augmented:} Kimberly \textcolor{violet}{travelled to club six times last month.} She buys 9 peanuts every time she goes to the club . \textcolor{violet}{How many peanuts did Kimberly buy last year ?}
\\ \cmidrule(l){2-3} 
 & \multirow{2}{*}{\textbf{\textcolor{red}{Problem Reordering}}} & \textbf{Original:} Fred has 10 blue marbles . Fred has number1 times more blue marbles than Tim . How many blue marbles does Tim have ?
\\
 &  & \textbf{Augmented:} \textcolor{violet}{If} fred has 10 blue marbles and fred has number1 more blue marbles \textcolor{violet}{than Tim then how many blue marbles does tim have ?}
 \\ \hline 
\multirow{6}{*}{\textbf{\begin{tabular}[c]{@{}l@{}}Substitution\\ Methods\end{tabular}}} & \multirow{2}{*}{\textbf{\textcolor{red}{Fill Masking}}} & \textbf{Original:} Sarah had 7 homework problems . She finished 2 of them but still had 3 pages of problems to do . If each page has the same number of problems on it , how many problems are on each page ?
\\
 &  & \textbf{Augmented:} Sarah had 7 of \textcolor{violet}{them} . She had 2 of \textcolor{violet}{them} but still had 3 more of \textcolor{violet}{them} to do . If each more has the same number of \textcolor{violet}{them} on it, \textcolor{violet}{How many them are on each more ?}
 \\ \cmidrule(l){2-3} 
 & \multirow{2}{*}{\textbf{\textcolor{red}{Named-Entity Replacement}}} & \textbf{Original:} Beverly had 10 dimes in his bank . His sister Maria borrowed 2 of his dimes . How many dimes does Beverly have now ?
\\
 &  & \textbf{Augmented:} \textcolor{violet}{Silva} had 10 dimes in his bank . His sister \textcolor{violet}{Jeanie} borrowed a \textcolor{violet}{pair} of his dimes . \textcolor{violet}{How many dimes does Jeanie have now ?} \\ \cmidrule(l){2-3} 
 & \multirow{2}{*}{\textbf{\textcolor{red}{Synonym Replacement}}} & \textbf{Original:} Shawn's team won their dodgeball game and scored 25 points total . If Shawn scored 13 of the points and everyone else scored 4 points each , how many players were on his team ?
 \\
 &  & \textbf{Augmented:} Shawn's \textcolor{violet}{group} won their \textcolor{violet}{rumble} game and scored 25 points total . If Shawn scored 13 of the points and everyone else scored \textcolor{violet}{quarter} points each, \textcolor{violet}{how many people were there ?}
\\
 \bottomrule
\end{tabular}
\caption{Invalid Augmentation examples from all proposed methods. Coloured text represents the changes in problem statement.}
\label{table:a3}
\end{table*}
\end{document}